\documentclass[]{esannV2}
\usepackage[a4paper]{geometry}
\usepackage[]{graphicx}
\usepackage{caption}
\usepackage{subcaption}
\usepackage[latin1]{inputenc}
\usepackage{amssymb,amsmath,array}
\usepackage{booktabs}
\usepackage{tabularx}
\usepackage{arydshln}  
\usepackage[binary-units=true]{siunitx}
\usepackage{tikz}
\usepackage{xcolor}
\usetikzlibrary{arrows,calc,math,patterns,angles,quotes,patterns,positioning,fit,backgrounds,angles,quotes}

\newcommand\Tstrut{\rule{0pt}{2.6ex}}         
\newcommand\Bstrut{\rule[-0.9ex]{0pt}{0pt}}   

\definecolor{unired}{RGB}{165,30,55} 
\definecolor{gray1}{RGB}{180, 180, 180}
\definecolor{redcustom}{RGB}{205, 0, 0}  
\definecolor{cobalt}{rgb}{0.0, 0.28, 0.67}

\usepackage[hidelinks]{hyperref}

\begin{document}
\title{%
A Distributed Neural Network Architecture for Robust Non-Linear Spatio-Temporal Prediction
}

\author{Matthias Karlbauer$^1$, Sebastian Otte$^1$,\\
	Hendrik P.A. Lensch$^3$, Thomas Scholten$^2$, Volker Wulfmeyer$^4$,\\
	 and Martin V. Butz$^1$
%
%
\vspace{.3cm}\\
%
1- University of T\"ubingen - Neuro-Cognitive Modeling Group \\
Sand 14, 72076 T\"ubingen - Germany
%
\vspace{.1cm}\\
2- University of T\"ubingen - Soil Science and Geomorphology \\
Rümelinstraße 19-23, 72070 T\"ubingen - Germany
\vspace{.1cm}\\
3- University of T\"ubingen - Computer Graphics \\
Maria-von-Linden-Straße 6, 72076 T\"ubingen - Germany
\vspace{.1cm}\\
4- University of Hohenheim - Institute for Physics and Meteorology  \\
Garbenstraße 30, 70599 Stuttgart - Germany\\
}

\maketitle


\begin{abstract}

We introduce a distributed spatio-temporal artificial neural network architecture (DISTANA).
It encodes mesh nodes using recurrent, neural prediction kernels (PKs), while neural transition kernels (TKs) transfer information between neighboring PKs, together modeling and predicting spatio-temporal time series dynamics. 
As a consequence, DISTANA assumes that generally applicable causes, which may be locally modified, generate the observed data. 
DISTANA learns in a parallel, spatially distributed manner, scales to large problem spaces, is capable of approximating complex dynamics, and is particularly robust to overfitting when compared to other competitive ANN models.
Moreover, it is applicable to heterogeneously structured meshes. 


\end{abstract}


\section{Introduction}

Modeling and predicting non-linear spatio-temporal process dynamics is challenging for current pattern recognition systems \cite{bronstein2017geometric}.
Representative problem domains range from the analysis of dynamic brain activities in neuroscience \cite{kasabov2014neucube}, over video streams \cite{karpathy2014large}, information flow in social networks \cite{de2018exploratory}, and traffic predictions \cite{zhao2017lstm}, to climate and weather forecasts \cite{Liu:2015,Shi:2015}.
In all cases, the major challenge is to infer, model, and predict the underlying causes 
that generate the perceived data stream, propagating the involved causal dynamics through graphs and distributed sensor meshes.
A key property, which all spatio-temporal processes have in common, is that some generally underlying principles---such as physics when observing natural processes---will apply irrespective of time or location. 
As a result, the same predictable patterns---individually modified by local spatial and temporal influences---will be observable repeatedly at different spatial locations and points in time.

We introduce DISTANA, a distributed spatio-temporal artificial neural network architecture, which actively searches for such characteristics in time series data.
DISTANA learns \emph{predictive}, spatio-temporal, neural network \emph{kernels} (PKs), which are applied to all nodes of a mesh. 
Additional information routing \emph{transition kernels} (TKs) laterally connect the PKs. 
Both PKs and TKs, respectively, share their weights, thus applying the same operations at different locations.
This enables efficient parallel computation in and learning from all spatial locations in the mesh. 
Moreover, DISTANA is predisposed to identify the universal, recurring causes of the observed pattern dynamics.
Compared to seven other ANN models, including convolutional neural networks (CNNs), recurrent neural networks (RNNs), and combinations of both (e.g. ConvLSTM), DISTANA reaches both higher accuracy and robustness at approximating circularly propagating waves, it is critically less prone to overfitting, and it bears the potential to handle heterogeneously distributed sensor meshes.
Thus, in the near future we intend to apply DISTANA to related, but more challenging real-world problems, such as learning to predict the partially chaotic processes that generate our weather and climate.


\section{Related Work}
While CNNs \cite{lecun1989backpropagation} have been shown to efficiently and accurately process spatially distributed information such as images, RNNs---and long short-term memory cells (LSTMs) \cite{hochreiter1997long} in particular---were designed to handle temporally distributed data such as time series.
Recently, Shi et al. \cite{Shi:2015} proposed ConvLSTM, a  combination of CNNs and LSTMs resulting in a convolution-gating architecture, which processes spatial and temporal information simultaneously.
GridLSTM \cite{kalchbrenner2015grid}, on the other hand, extends LSTMs to process not only temporal but also spatial data dimensions sequentially.
DISTANA belongs to a third related class of architectures, which is referred to as graph neural networks (GNNs) \cite{scarselli2008graph}. 
GNNs treat graph vertices and edges in two different neural network components.
Unlike earlier GNNs, however, DISTANA integrates LSTM structures, projects the graph, i.e. its mesh, onto a metrical space, and assumes universal causes underlying the observable spatio-temporal data. 



\section{Model Description}

\begin{figure}[t]
	\centering
	\begin{minipage}{0.38\textwidth}
		\centering
		\includegraphics[width=\textwidth]{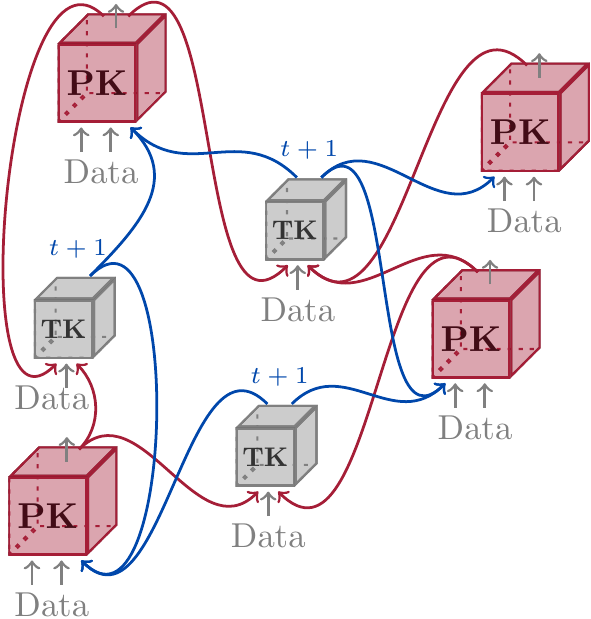}
	\end{minipage}
	\hfill
	\begin{minipage}{0.58\textwidth}
		\centering
		\includegraphics[width=\textwidth]{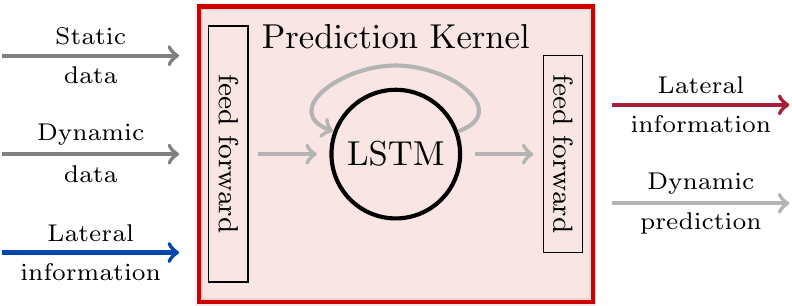}\\
		\vspace{0.1cm}
		\includegraphics[width=\textwidth]{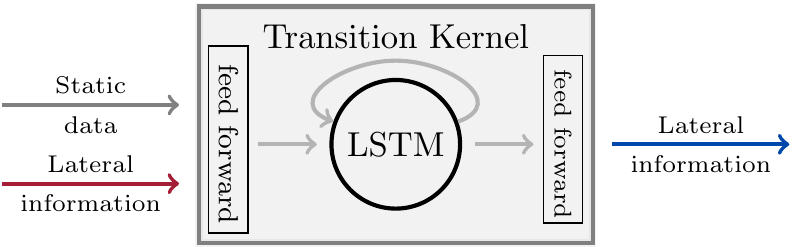}
	\end{minipage}
	\caption{Left: the lateral connection schema of Prediction- and Transition Kernels. Right: the inputs, outputs and an exemplary topology of Prediction- and Transition Kernels with recurrent connections.}\label{fig:network_architecture}
\end{figure}

DISTANA is a two-network architecture that consists of a PK network, which generates dynamic predictions at each desired spatial position, and a TK network, which models transitions between (two or more) adjacent PKs. 
The PKs and the TKs, which share their respective weights, are applied in a sensor mesh. 
As depicted in \autoref{fig:network_architecture}, the PK and TK networks can be applied simultaneously in space, processing spatially distributed data.
Each PK instance receives (1) dynamic input, which is subject to prediction and changes over time, (2) static information, which stays constant and characterizes the location of each PK, and (3) lateral input from neighboring PKs.
The TK network---making our approach unique---is introduced to model location-sensitive transitions between PKs and thus to enable local context-dependent spatial information propagation.
In principle, both PK and TK networks can have arbitrary topologies and may incorporate recurrent connections (cf. \autoref{fig:network_architecture}).


\section{Experiments}
\begin{figure}
	\centering
	\includegraphics[width=.8\textwidth]{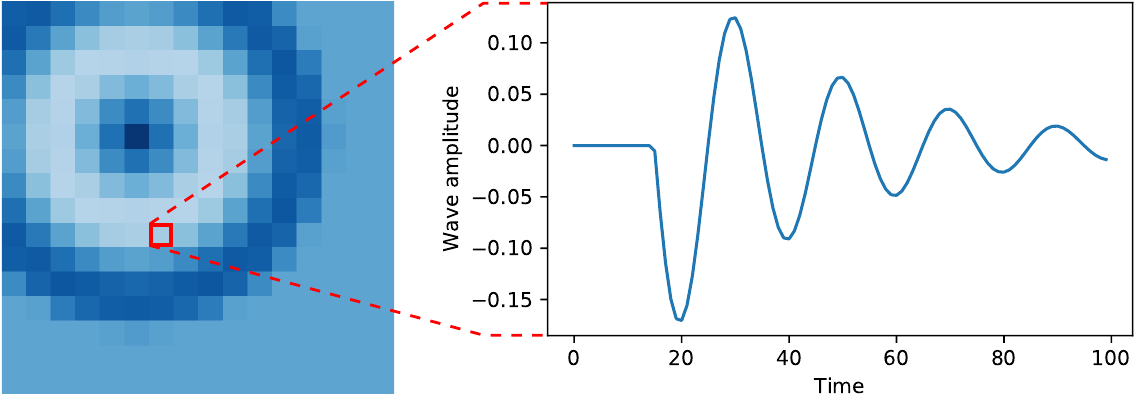}
	\caption{Data set one. Left: exemplar circular wave. Right: activity pattern over time at one particular position in the two-dimensional wave field.}\label{fig:circular_wave}
\end{figure}

In two experiments, which differ in the data sets used, several ANN architectures including fully connected networks, CNNs, and RNNs are compared with DISTANA, 
modeling a wave-like spatio-temporal process (cf. \autoref{fig:circular_wave}), which is distributed in a $16 \times 16$ mesh.
Train and test errors are calculated as mean squared errors between network output and target, being the network input shifted by one time step, requiring the networks to predict the next time step of a 2D circular wave sequence.
The test error is calculated over 65 time steps of closed loop performance, where the network feeds itself with its own dynamic predictions from the previous time step.
The closed loop begins after 15 steps of teacher forcing, which ground the recurrent activity in the network.

\subsection{Data Set 1}
Initially, a basic data set is created where single waves are generated propagating outwards. 
The waves are not reflected at the borders, yielding comparably simple dynamics.
Waves were generated using
\begin{equation}
	u(x, y, t) = \begin{cases}\sin(r_{x, y} - ct)\exp(-d(ct - r_{x, y})) & \text{if }r_{x, y} < ct\\
	0 & \text{else}
	\end{cases},
\end{equation}\label{eq:simple_wave_generation}%
where $u(x, y, t)$ is the wave height of the field at a certain position and time, $\sin(r_{x, y} - ct)$ defines the oscillating wave height considering the distance to the wave center $r_{x, y}$ and the current time step $t$, and $\exp(-d(ct - r_{x, y}))$ can be described as decaying expression, which makes waves decay with respect to their distance to the wave center over time while considering a decay factor $d = 0.25$.
For large values of $d$ the wave fades quicker than for small values of $d$.
Constant $c = 10$ is the wave velocity and field values which have not been reached by the wave in time step $t$ are explicitly set to zero.

\begin{table}[b!]
	\centering
	\caption{Average prediction error measured over 65 closed-loop time steps, number of parameters, and inference time for different ANNs, including DISTANA, approximating circular wave propagations.}\label{tab:architecture_performances}
    \small
		\begin{tabularx}{\linewidth}{Xrrrrr}
	\toprule
	Model (\#pars) & Train error & Test error & Inf. time & 1-train-ex. & Var. wave\\
	\midrule
	Baseline $t-1$ & - & \SI{3.59e-5}{} & - & \SI{3.59e-5}{} & \SI{5.90e-5}{}\\
	Baseline zero & - & \SI{8.88e-5}{} & - & \SI{8.88e-5}{} & \SI{5.84e-4}{}\Bstrut\\\hdashline\Tstrut
	FC-Linear (65k) & \SI{2.36e-4}{} & \SI{2.56e-4}{} & \bfseries{\SI{0.0198}{\s}} & \SI{2.41e-4}{} & \SI{1.91e-3}{}\\				
	FC-LSTM (524k) & \SI{5.34e-5}{} & \SI{1.38e-2}{} & \SI{0.0279}{\s} & \SI{2.14e-3}{} & \SI{6.57e-3}{}\Bstrut\\\hdashline\Tstrut
	CNN (20) & \SI{3.41e-4}{} & \SI{2.22e-4}{} & \SI{0.0479}{\s} & \SI{1.37e-3}{} & \SI{2.66e-2}{}\\
	TCN (2.3k) & \SI{1.17e-5}{} & \SI{8.56e-3}{} & \SI{0.0787}{\s} & \SI{1.04e-1}{} & \SI{3.09e-2}{}\Bstrut\\\hdashline\Tstrut
	CLSTMC (768k) & \SI{6.28e-5}{} & \SI{4.67e-1}{} & \SI{0.0683}{\s} & \SI{6.13e-4}{} & \SI{2.71e-1}{}\\
	ConvLSTM1 (144) & \SI{1.83e-5}{} & \SI{4.26e-5}{} & \SI{0.0931}{\s} & \SI{4.55e-5}{} & \SI{5.85e-4}{}\\
	ConvLSTM8 (2.9k) & \SI{6.34e-6}{} & \bfseries{\SI{1.28e-6}{}} & \SI{0.0959}{\s} & \SI{1.29e-2}{} & \SI{7.88e-4}{}\Bstrut\\\hdashline\Tstrut
	GridLSTM (624) & \SI{7.95e-5}{} & \SI{3.62e-1}{} & \SI{5.8786}{\s} & \SI{2.86e-1}{} & \SI{1.35e-1}{}\\
	BiGridLSTM (1.8k) & \bfseries{\SI{6.28e-06}{}} & \SI{5.65e-1}{} & \SI{11.9900}{\s} & \SI{8.67e-1}{} & \SI{4.55e-2}{}\Bstrut\\\hdashline\Tstrut
	DISTANA4 (106) & \SI{2.32e-5}{} & \SI{9.54e-6}{} & \SI{0.0821}{\s} & \SI{2.29e-5}{} & \SI{2.17e-4}{}\\
	DISTANA26 (2.9k) & \SI{1.34e-5}{} & \SI{3.18e-5}{} & \SI{0.1022}{\s} & \bfseries{\SI{2.12e-5}{}} & \bfseries{\SI{4.70e-05}{}}\\
	\bottomrule
\end{tabularx}

\end{table}

\begin{figure}[b]
	\centering
	\includegraphics[height=0.23\textwidth]{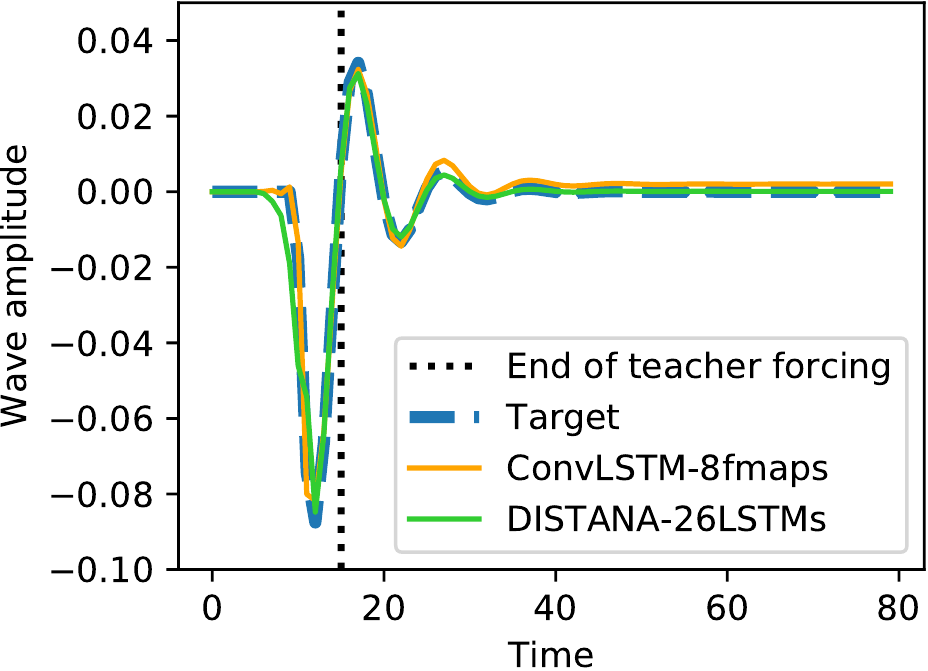}
	\hfill
	\includegraphics[height=0.23\textwidth]{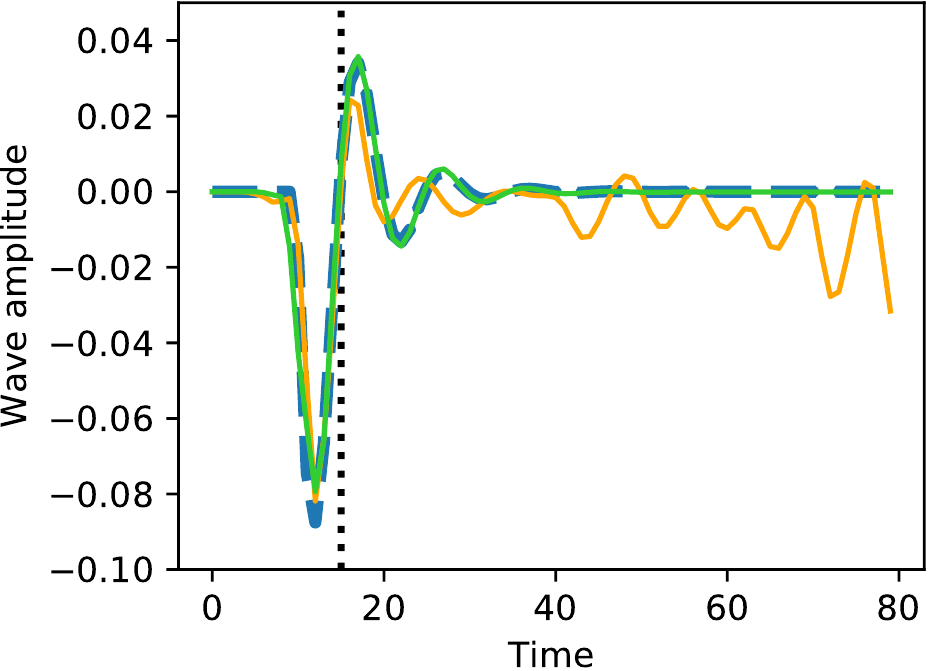}
	\hfill
	\includegraphics[height=0.23\textwidth]{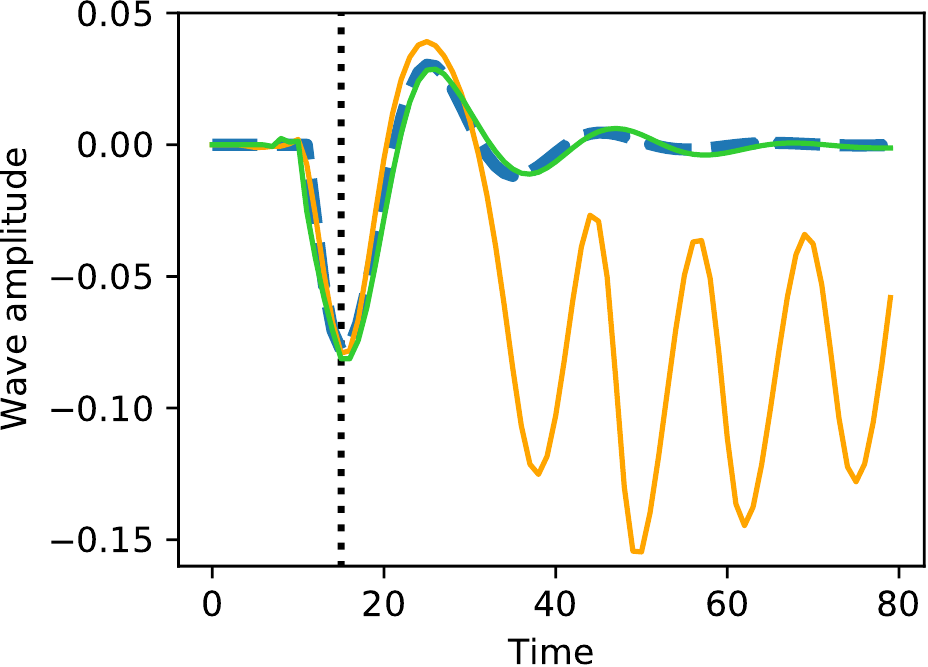}
	\caption{Network dynamics generated by the most promising ANN architectures.
		Left: normal training; center: training on one single wave sequence (i.e. 1-train-ex); right: training on waves with variable velocities (i.e. Var. wave).}\label{fig:netouts}
\end{figure}

\autoref{tab:architecture_performances} shows the performance of all compared models at approximating the circular wave.
Additionally to the \emph{train} and \emph{test error}, we report the number of \emph{parameters} and the \emph{inference time} of one sequence (consisting of 80 time steps) for each model.
In order to rigorously test all models for their generalization abilities, we also trained them on one single sequence and again computed the test error on unseen sequences (test error \emph{1-train-ex.}).
Furthermore, to elaborate the models' abilities to approximate variable dynamics, we trained them on waves that travel with varying velocities (test error \emph{var. wave}).
See \autoref{fig:netouts} for a performance visualization.
Spatial scalability, which indicates whether a model can be applied to an input field of different resolution, is reported in the subsequent model descriptions.

\paragraph{Baselines} Two baselines were created as upper bounds. \emph{Baseline $t-1$} was calculated by assuming an identity function that returns the input directly, whereas \emph{Baseline zero} assumes a model that always predicts zeros.

\paragraph{Fully Connected Networks} A naive and spatially not scalable approach to model the circular wave is a fully connected linear network (\emph{FC-Linear}), with $16 \times 16 = 256$ cells, receiving the flattened input.
A more elaborated model is \emph{FC-LSTM}, which replaces the linear layer of \emph{FC-Linear} by a 256-cell LSTM layer to facilitate temporal information processing.

\paragraph{CNN} To reduce the number of parameters, defining a spatially scalable model, numerous \emph{CNN}s with different kernel sizes, a varying number of feature maps, and two convolutional layers were evaluated.
The best results, which are reported here, were achieved by using a kernel size of $3\times 3$ and one feature map.

\paragraph{Temporal Convolution Network (TCN)} TCNs, as a spatially scalable approach, were applied with three 3D convolution layers, each with a $3\times 3\times 3$ kernel and $[1, 8, 1]$ feature maps.
Other depths or kernel sizes did not seem to improve performance.

\paragraph{CNN-LSTM-CNN (CLSTMC)} The \emph{CNN} approach was extended by inserting a fully connected LSTM layer---making it not spatially scalable---after a variable number of layers (one to three convolution layers followed by the LSTM layer and one to three transposed convolution layers).
Best results were achieved with one $3\times 3$ convolution followed by a flat LSTM layer and a $3\times 3$ transposed convolution with skip connection.

\paragraph{ConvLSTM} Two models of the spatially scalable ConvLSTM architecture, both with two layers and kernel size three, are reported: \emph{ConvLSTM1} with one feature map in both layers, and \emph{ConvLSTM8} with eight feature maps in the first layer, which are reduced to one in the second layer.

\paragraph{GridLSTM and BiGridLSTM} Spatially scalable GridLSTM models are evaluated. \emph{GridLSTM} runs forward over all three data dimensions. \emph{BiGridLSTM} processes the data forward in time but bidirectionally over space.

\paragraph{Our model (DISTANA)} The PK consists of a linear layer, followed by a layer of either four or 26 LSTM cells, and another linear layer.
The TK is a simple linear layer and is used---like the other linear layers---without activation function.
As some of the other models above, DISTANA is spatially scalable.

\subsection{Data Set 2}
To increase data complexity, a second set was created where waves are reflected at borders, such that wave fronts become interactive.
We focus our analysis on the most promising architectures determined above. 
For wave data generation, the following two-dimensional wave equation:
\begin{equation}
	\frac{\partial^2 u}{\partial t^2}=c^2\left(\frac{\partial^2 u}{\partial x^2} + \frac{\partial^2 u}{\partial y^2}\right)\label{eq:wave_equation}
\end{equation}%
was solved numerically using the second order central differences approach
\begin{equation}
	\frac{\partial^2 u}{\partial b^2} =\frac{u(b+h)-2u(b)+u(b-h)}{h^2} = u_{bb},\label{eq:second_order_central_differences}
\end{equation}%
where $b$ stands for a variable of function $u$ and $h$ is the approximation step size with respect to the considered variable $b$.

When applying \autoref{eq:second_order_central_differences} to \autoref{eq:wave_equation}, we obtain
\begin{equation}
	\frac{u(x, y, t+\Delta_t) - 2u(x, y, t) +u(x, y, t-\Delta_t)}{\Delta_t^2} = c^2(u_{xx} + u_{yy})
\end{equation}%
which can be solved for $u(x, y, t+\Delta_t)$ to obtain an equation for computing the state of the field at a desired position in the subsequent time step $t+\Delta_t$
\begin{equation}
	u(x, y, t+\Delta_t) = c^2\Delta_t^2(u_{xx} + u_{yy}) + 2u(x, y, t) - u(x, y, t-\Delta_t).
\end{equation}%
Boundary conditions in both space (when $x < 0$ or $x > \text{field width}$, analogously for $y$) and time domains (in time step 0) are treated as zero.
The following variable choices were met: $\Delta_t = 0.1$, $\Delta_x = \Delta_y= 1$ and $c = 3.0$.
The field was initialized using the Gaussian bell curve
\begin{equation}
	u(x, y, 0) = a\exp\left(-\left(\frac{(x - s_x)^2}{2\sigma^2_x} + \frac{(y - s_y)^2}{2\sigma^2_y}\right)\right)
\end{equation}
with amplitude factor $a = 0.34$, wave width in $x$- and $y$-direction $\sigma^2_x = \sigma^2_y = 0.5$ and $s_x, s_y$ being the starting point or center of the circular wave.

\begin{figure}
	\centering
	\includegraphics[height=0.31\textwidth]{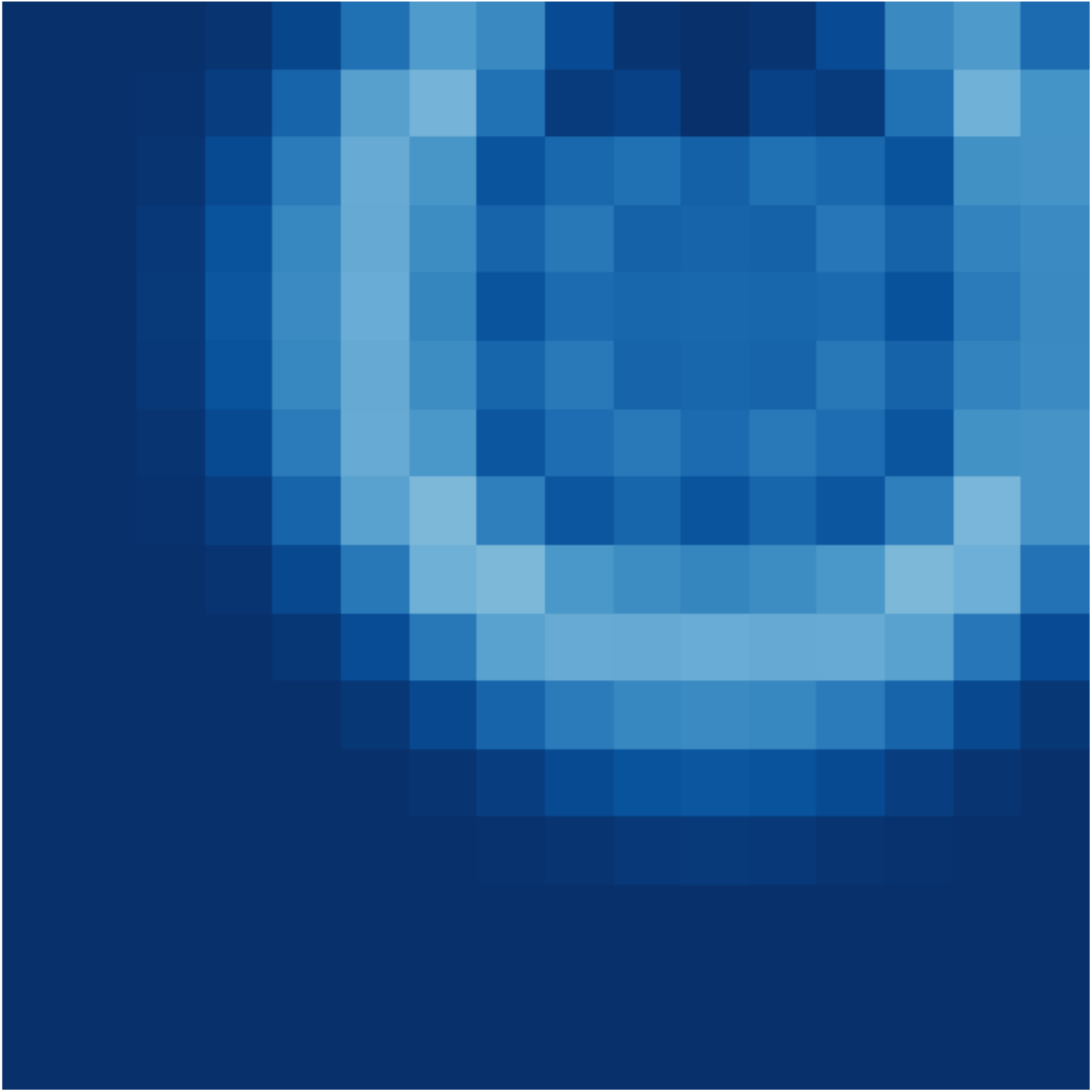}
	\hfill
	\includegraphics[height=0.31\textwidth]{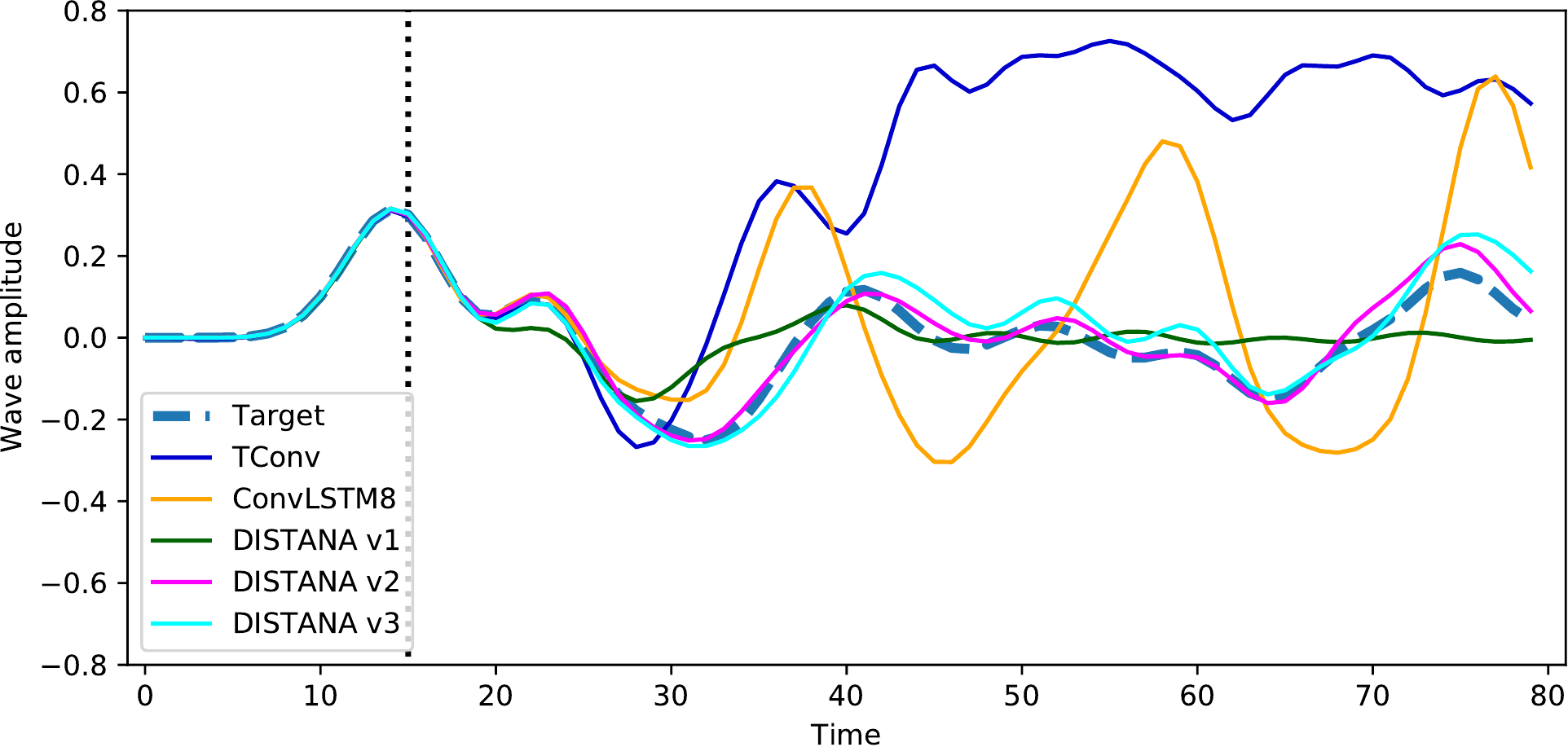}
	\caption{Data set two. Left: exemplary circular wave with reflecting borders. Right: network dynamics generated by selected architectures.}\label{fig:netouts_new_data}
\end{figure}

The unfolding dynamics of higher complexity are much harder to predict, as can be seen in \autoref{fig:netouts_new_data}.
None of the previously tested architectures was able to approximate the dynamics satisfactorily, as can be seen in the test error rates of \autoref{tab:new_data_comparison} remaining larger than the baseline test errors. 
Accordingly, DISTANA was adapted slightly in three ways, described in the following.

\begin{table}[b]
	\centering
	\caption{Same evaluation as in \autoref{tab:architecture_performances} on data set two, including TCN, ConvLSTM and three variants of DISTANA.}\label{tab:new_data_comparison}
	\small
	\begin{tabularx}{\linewidth}{lXrXrXr}    
	\toprule
	Model (\#pars) & \ & Train error & \ & Test error & \ & Inf. time\\
	\midrule
	Baseline $t-1$ & & - & & \SI{5.83e-3}{} & & -\\
	Baseline zero & & - & & \SI{1.07e-2}{} & & -\Bstrut\\\hdashline\Tstrut
	TCN (2.3k) & & \SI{1.14e-5}{} & & \SI{2.11e-1}{} & & \SI{0.0625}{\s}\Bstrut\\\hdashline\Tstrut
	ConvLSTM8 (2.9k) & & \SI{3.53e-6}{} & & \SI{2.35e-2}{} & & \SI{0.0461}{\s}\Bstrut\\\hdashline\Tstrut
	DISTANAv1 (146) & & \SI{5.27e-6}{} & & \SI{9.10e-3}{} & & \SI{0.0338}{\s}\\
	DISTANAv2 (172) & & \bfseries{\SI{1.88e-6}{}} & & \bfseries{\SI{4.60e-4}{}} & & \SI{0.0316}{\s}\\
	DISTANAv3 (200) & & \SI{3.60e-6}{} & & \SI{3.55e-3}{} & & \bfseries{\SI{0.0233}{\s}}\\
	\bottomrule
\end{tabularx}
\end{table}

\paragraph{DISTANA v1} The size of the preprocessing feed forward layer in the PK was increased from one to four neurons.

\paragraph{DISTANA v2} Next to an increased preprocessing layer as in \emph{DISTANA v1}, the lateral input neurons were changed from one to eight.
Furthermore, instead of using TKs as information transition tools, the WKs were designed to communicate with all eight grid neighbors directly.
Each lateral input neuron consistently receives input from a particular neighbor, depending on the direction.

\paragraph{DISTANA v3} While the same changes as in \emph{DISTANA v2} were applied, the number of lateral output neurons was increased from one to eight, analogously to the lateral input neurons.

As a result, DISTANAv2 and DISTANAv3 strongly outperform the simpler DISTANA version as well as TCN and ConvLSTM. Table~\ref{tab:new_data_comparison} shows that DISTANAv2 not only reaches the lowest training error but also yields the best generalization performance. 
Fig.~\ref{fig:netouts_new_data} shows that when closed loop predictions unfold after 15 steps of teacher forcing, DISTANAv2 and DISTANAv3 approximate the target value still rather well while the other ANN architectures start to strongly deviate from the target values after only 10 to 15 closed-loop prediction steps. 
\href{https://www.youtube.com/watch?v=4VHhHYeWTzo}{Online video material}\footnote{\href{https://www.youtube.com/watch?v=4VHhHYeWTzo}{https://www.youtube.com/watch?v=4VHhHYeWTzo}} illustratively shows the further abilities of DISTANA, including its ability to generalize to larger grid sizes.


\section{Discussion}

Several ANN architectures were compared at approximating a spatio-temporal process, the circular wave, using two different complexity scenarios.
The performance comparisons of data set one show that only ConvLSTM and our model, DISTANA, yield smaller test errors than the two baselines.
Recall that here the closed loop performance over $T$ prediction time steps was measured, which is much harder than just next time step prediction, as it requires both intrinsic model stability and the maintenance of plausible ongoing dynamics.
While the reported accuracy for standard training and simple dynamics is in favor of ConvLSTM, DISTANA proved robust to few and variable training data (see \autoref{fig:netouts} and last two columns in \autoref{tab:architecture_performances}), even with a network that contains only 106 parameters.
In these latter cases, \emph{ConvLSTM1} tended to approach the zero baseline, while DISTANA still predicted the actual wave amplitude, albeit with a slightly differing wave frequency, which prevented an even smaller error.
\emph{ConvLSTM8} reached outstanding test errors but tended to overfit heavily, as visible by the comparably bad testing error when trained on one training example only (\autoref{fig:netouts}).
These findings were corroborated by the evaluations in a second, more complex data set, in which waves were reflected at borders and thus heavily interacted with each other. 
All other considered architectures failed to generate lasting closed-loop predictions, except for two variants of DISTANA that considered lateral information propagation explicitly (\autoref{fig:netouts_new_data}).
DISTANA did not tend to overfit and generalized very well, because it assumes equal dynamic propagation principles throughout space and time by design.  

So far, we have only considered regularly distributed grid data, where distances between single measurement points are identical.
However, this data situation is not given in many applications, such as neural information processing in the brain, traffic predictions on roads, data propagation on graphs, or weather prediction given weather station network, radar, and satellite data.
In all these cases, distances and the number of neighboring vertices vary and we are currently enhancing the TKs in DISTANA accordingly. 
Ongoing work shows that DISTANA can indeed handle irregularly distributed sensor meshes.
We are particularly interested in whether DISTANA will detect coherent and predictable structures in weather data, which will indicate its scalable applicability to short-range weather forecasting.



\subsection*{Acknowledgments}
We thank Georg Martius for inspiring ideas and Nicholas Kr\"{a}mer for sharing expertise in complex data generation; we thank the International Max Planck Research School for Intelligent Systems (IMPRS-IS) for supporting Matthias Karlbauer, and gratefully mention that this work is funded by the German Research Foundation (DFG) under Germany's Excellence Strategy -- EXC-Number 	2064/1 -- Project Number 390727645.


\begin{footnotesize}

\bibliographystyle{unsrt}
\bibliography{literature}

\begin{thebibliography}{10}

\bibitem{bronstein2017geometric}
Michael~M Bronstein, Joan Bruna, Yann LeCun, Arthur Szlam, and Pierre
  Vandergheynst.
\newblock Geometric deep learning: going beyond euclidean data.
\newblock {\em IEEE Signal Processing Magazine}, 34(4):18--42, 2017.

\bibitem{kasabov2014neucube}
Nikola~K Kasabov.
\newblock Neucube: A spiking neural network architecture for mapping, learning
  and understanding of spatio-temporal brain data.
\newblock {\em Neural Networks}, 52:62--76, 2014.

\bibitem{karpathy2014large}
Andrej Karpathy, George Toderici, Sanketh Shetty, Thomas Leung, Rahul
  Sukthankar, and Li~Fei-Fei.
\newblock Large-scale video classification with convolutional neural networks.
\newblock In {\em Proceedings of the IEEE conference on Computer Vision and
  Pattern Recognition}, pages 1725--1732, 2014.

\bibitem{de2018exploratory}
Wouter De~Nooy, Andrej Mrvar, and Vladimir Batagelj.
\newblock {\em Exploratory social network analysis with Pajek: Revised and
  expanded edition for updated software}, volume~46.
\newblock Cambridge University Press, 2018.

\bibitem{zhao2017lstm}
Zheng Zhao, Weihai Chen, Xingming Wu, Peter~CY Chen, and Jingmeng Liu.
\newblock Lstm network: a deep learning approach for short-term traffic
  forecast.
\newblock {\em IET Intelligent Transport Systems}, 11(2):68--75, 2017.

\bibitem{Liu:2015}
J.~N.~K. Liu, Y.~Hu, Y.~He, P.~W. Chan, and L.~Lai.
\newblock {\em Information Granularity, Big Data, and Computational
  Intelligence}, volume~8 of {\em Studies in Big Data}, chapter Deep Neural
  Network Modeling for Big Data Weather Forecasting, pages 389--408.
\newblock Springer, 2015.

\bibitem{Shi:2015}
Xingjian Shi, Zhourong Chen, Hao Wang, Dit-Yan Yeung, Wai-kin Wong, and
  Wang-chun WOO.
\newblock Convolutional {LSTM} network: A machine learning approach for
  precipitation nowcasting.
\newblock In C.~Cortes, N.~D. Lawrence, D.~D. Lee, M.~Sugiyama, and R.~Garnett,
  editors, {\em Advances in Neural Information Processing Systems 28}, pages
  802--810. Curran Associates, Inc., 2015.

\bibitem{lecun1989backpropagation}
Yann LeCun, Bernhard Boser, John~S Denker, Donnie Henderson, Richard~E Howard,
  Wayne Hubbard, and Lawrence~D Jackel.
\newblock Backpropagation applied to handwritten zip code recognition.
\newblock {\em Neural computation}, 1(4):541--551, 1989.

\bibitem{hochreiter1997long}
Sepp Hochreiter and J{\"u}rgen Schmidhuber.
\newblock Long short-term memory.
\newblock {\em Neural computation}, 9(8):1735--1780, 1997.

\bibitem{kalchbrenner2015grid}
Nal Kalchbrenner, Ivo Danihelka, and Alex Graves.
\newblock Grid long short-term memory.
\newblock {\em arXiv preprint arXiv:1507.01526}, 2015.

\bibitem{scarselli2008graph}
Franco Scarselli, Marco Gori, Ah~Chung Tsoi, Markus Hagenbuchner, and Gabriele
  Monfardini.
\newblock The graph neural network model.
\newblock {\em IEEE Transactions on Neural Networks}, 20(1):61--80, 2008.

\end{thebibliography}

\end{footnotesize}


\end{document}